\documentclass[conference]{IEEEtran}
\IEEEoverridecommandlockouts

\usepackage{fancyhdr}
\usepackage{cite}
\usepackage{amsmath,amssymb,amsfonts}
\usepackage{algorithmic}
\usepackage{graphicx}
\usepackage{textcomp}
\usepackage{subcaption}
\usepackage{multirow}
\usepackage{pbox}
\usepackage{multicol}

\def\BibTeX{{\rm B\kern-.05em{\sc i\kern-.025em b}\kern-.08em
    T\kern-.1667em\lower.7ex\hbox{E}\kern-.125emX}}

\DeclareMathOperator{\score}{score}

\DeclareMathOperator{\va}{\mathbf a}

\DeclareMathOperator{\vc}{\mathbf c}

\DeclareMathOperator{\vh}{\mathbf h}

\DeclareMathOperator{\vx}{\mathbf x}
\DeclareMathOperator{\vy}{\mathbf y}

\DeclareMathOperator{\vX}{\mathbf X}

\begin{document}

\title{Context-aware Cascade Attention-based RNN for Video Emotion Recognition\\
{\footnotesize \textsuperscript{}}
\thanks{}
}

\author{\IEEEauthorblockN{Man-Chin Sun}
\IEEEauthorblockA{
\textit{Emotibot Inc.}\\
Taipei, Taiwan \\
manchinsun@emotibot.com}
\and
\IEEEauthorblockN{Shih-Huan Hsu}
\IEEEauthorblockA{
\textit{Emotibot Inc.}\\
Taipei, Taiwan \\
cyrilhsu@emotibot.com}
\and
\IEEEauthorblockN{Min-Chun Yang}
\IEEEauthorblockA{
\textit{Emotibot Inc.}\\
Taipei, Taiwan \\
kaiyang@emotibot.com}
\and
\IEEEauthorblockN{Jen-Hsien Chien}
\IEEEauthorblockA{
\textit{Emotibot Inc.}\\
Taipei, Taiwan \\
kennychien@emotibot.com}
}

\maketitle

\begin{abstract}
Emotion recognition can provide crucial information about the user in many applications when building human-computer interaction (HCI) systems. Most of current researches on visual emotion recognition are focusing on exploring facial features. However, context information including surrounding environment and human body can also provide extra clues to recognize emotion more accurately. Inspired by ``sequence to sequence model'' for neural machine translation, which models input and output sequences by an encoder and a decoder in recurrent neural network (RNN) architecture respectively, a novel architecture, ``CACA-RNN'', is proposed in this work. The proposed network consists of two RNNs in a cascaded architecture to process both context and facial information to perform video emotion classification. Results of the model were submitted to video emotion recognition sub-challenge in Multimodal Emotion Recognition Challenge (MEC2017). CACA-RNN outperforms the MEC2017 baseline (mAP of 21.7\%): it achieved mAP of 45.51\% on the testing set in the video only challenge.
\end{abstract}


\section{Introduction}
Understanding human emotion has attracted a lot of attention recently. And it also plays an important role in many applications such as human-computer interaction, advertising, social media communication and cognitive science. However, emotion recognition is still a challenging task. It's very difficult to find an effective model for emotion and facial expressions, let alone combining of multimodal data of visual, vocal and even text to emotion recognition. In this work, a novel model is proposed to consider facial and context information concurrently, which leads to superior performance in emotion recognition.

One of the key problems of emotion recognition is emotion representation or emotion model. There are many researches about emotion representation, such as six discrete basic emotion classes proposed by Ekman \cite{Ekman1971_cultures}, continuous dimensional models (e.g. valence and arousal in \cite{Russell1980_circumplex}), and facial action coding system (FACS) in \cite{Ekman1977_facs}, which describes facial movement in action units (AU). Combination of facial action units can be classified into different emotions.

Many emotion datasets and challenges have also been published using aforementioned emotion representation models such as AFF-Wild \cite{Zafeiriou2017_aff}, which uses valence and arousal space to model facial expression. For the emotion classification representation, challenges such as Emotion Recognition Challenge in the Wild (EmotiW) \cite{Dhall2017_emotiw} and Multimodal Emotion Recognition Challenge (MEC) \cite{Li2017_mec}, which is the challenge of this work attending, have been held for recent years. Furthermore, some challenges such as \cite{Dhall2017_emotiw, Li2017_mec} also includes vocal features for emotion recognition.

In psychological researches \cite{hess2015_context, Barrett2011_context}, it has been discussed that other than facial expressions, contextual information such as body, pose and surrounding environment can also provide important clues for emotion perception. Evidence and experiments are also provided in \cite{hess2015_context, Barrett2011_context} to show that emotion perception can be influenced by context. Moreover, in some cases, context is even indispensable for emotion communication. Similar results are also proposed in computer vision literatures. Experiments in \cite{Kosti2017_emotion} show that when using both context and body information, performance of emotion recognition outperforms that of using only body image or only context image. A dataset ``Emotions in Context Database'' (EMOTIC) has also been published recently in \cite{Kosti2017_emotion}. 

Most of recent emotion recognition methods focus on exploring facial features based on deep neural network. Convolutional Neural Network (CNN) has been used to extract face features in some works \cite{Liu2014_Emotionet, Yao2016_HoloNet}. Some researches incorporate 3D Convolutional Networks (C3D) and Recurrent Neural Network (RNN) to model spatial and temporal clues of faces \cite{Pini2017_Multimodal, Fan2017_hybrid}. Also some works \cite{Fan2017_hybrid, Pini2017_Multimodal, Hu2017_ensemble} combine audio models to perform multimodal emotion recognition. 

This work focuses on video emotion classification using sequence of both facial and context information. A Context-Aware Cascade Attention-based RNN is proposed here to leverage both facial and context features to perform video emotion recognition. To further evaluate the influence of context information, multiple RNN-based networks are designed to compare different methods of fusing face and context features.

\section{Related Works}

Many works proposed video recognition methods based on neural networks. C3D networks were utilized to learn temporal information from sequential images for action recognition \cite{Karpathy2014_video, Tran2014_C3D} and video emotion recognition \cite{Fan2017_hybrid}. Recurrent neural network, which has temporal recurrence of latent variables, was proposed in \cite{Rumelhart1985_RNN}. Long-Short Term Memory (LSTM) was proposed in \cite{Hochreiter1997_LSTM} to handle long-range sequence learning.

``Sequence-to-sequence'' model \cite{Sutskever2014_seq}, which models sequence encoding and generation using LSTM RNNs, is one of the popular architectures of learning from sequential data. It has achieved the state-of-the-art results in neural machine translation. The sequence-to-sequence model consists of a RNN-based encoder and a decoder to learn temporal structures and generate output sequence. Furthermore, temporal attention mechanism has also been proposed in \cite{Bahdanau2015_nmt} as a soft-alignment method in machine translation to learn relevant temporal segments from encoder and decoder.

The encoder-decoder framework in sequence-to-sequence has also been used to describe videos with CNN and RNN to generate captions \cite{Sutskever2014_totext}. The temporal attention mechanism has also been incorporated in \cite{Yao2015_videos} to learn relevant temporal intervals from video sequences.

\section{Proposed Method}

In this section, the effect of context on improving emotion perception is first shown in Section. \ref{effect_context}. Then, inspired by Sequence-to-sequence model and attention mechanism for neural machine translation, a novel architecture for sequence classification will be proposed. Sequence-to-sequence model and attention mechanism will be briefly described in Section. \ref{seq2seq} and \ref{atten} respectively. Then the proposed model will be introduced in detail in Section. \ref{CACA-model}.

\begin{figure}[htb]
\centering
\includegraphics[width=0.7\columnwidth]{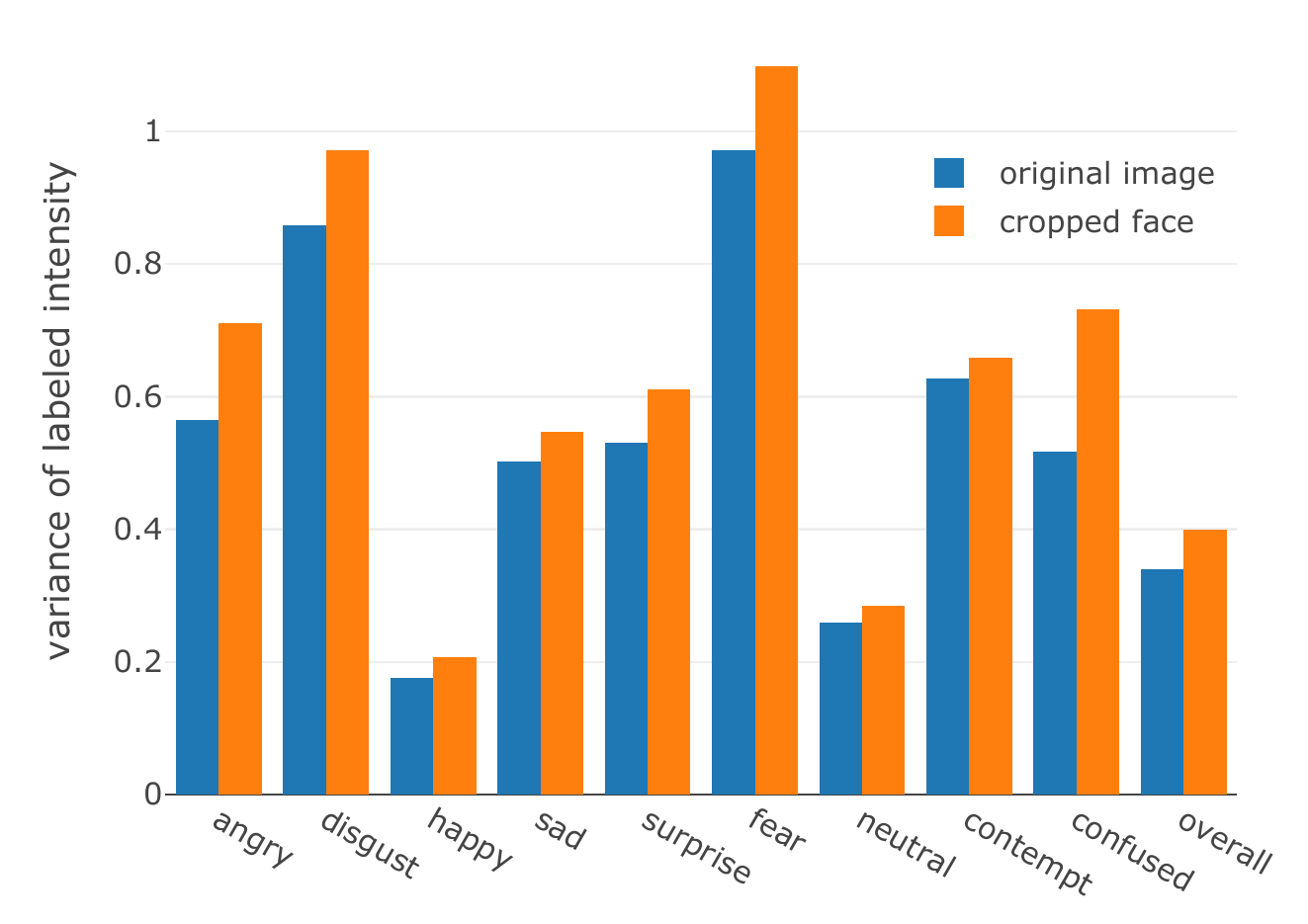}
\caption{Visualization of human emotion perception.\\
The dataset described in Section. \ref{effect_context} is labeled twice: 
\textbf{original images} are the label obtained providing the labelers with
original images; \textbf{cropped face} are the labels when only lablers are
provided with cropped face.
}
\label{fig-emtion-perception}
\end{figure}

\subsection{Effect of Context Information on Emotion Perception} \label{effect_context}

To visualize the effect of context information to emotion perception for human, an experiment was designed to show disagreement of human label of the same dataset with and without context information. The dataset consists of 55504 images with bonding box of each face. Each image is labeled by intensity of 9 emotions: ``angry'', ``disgust'', ``happy'', ``sad'', ``surprise'', ``fear'', ``neutral'', ``contempt'', ``confused'' and intensity is in score from 0 to 5 by each labeler. Each image is labeled by at least 5 labelers. Then the disagreement of each emotion is calculated by averaging the variance of labeled intensity of each image.

In Fig. \ref{fig-emtion-perception}, it is shown that when providing the labelers with original images of both face and context, more consensus can be observed in the obtained labels. Furthermore, emotion class ``happy'' and ``neutral'' have the least disagreement, which means happy and neutral can be recognized easily, while other emotions of higher disagreement shows more confusion in labelers.

\subsection{Sequence-to-Sequence Framework} \label{seq2seq}

Sequence-to-sequence, which is an encoder-decoder framework, was proposed in \cite{Sutskever2014_seq} to perform machine translation. The encoder reads a sequence of input vectors over input time $1\ldots T$ as $\vX=(\vx_1, \ldots, \vx_{T})$, into a ``context vector'' at encoder $\vc_{e,T}$. The context vector here represents the stored state and information of the encoder. When the encoder is a RNN, the hidden states are represented as $\vh_{e,t} = f(\vx_{e,t}, \vh_{e,t-1})$ and $\vc_{e,T}=g(\vh_{e,1}, \vh_{e,2}, \ldots, \vh_{e,T})$, where $f(\cdot)$ and $g(\cdot)$ are nonlinear functions. 

The decoder then generates one prediction at output time $i$, $\vy_i$, from context vector $\vc_{e, T}$ and all the previous predicted output. With a RNN decoder, its conditional probability models as 
\begin{align}
p(\vy_i | \vy_1, \ldots, \vy_{i-1}, \vc_{e,T}) = h(\vy_{i-1}, \vc_{e,T}, \vh_{d,i}),
\end{align}
where $\vh_{d,i}$ is the hidden states of the decoder RNN at output time $i$, $\vh_{d,i} = f(\vh_{d,i-1}, \vy_{i-1}, \vc_{e,T})$, and $f(\cdot)$, $h(\cdot)$ are nonlinear functions.

\begin{figure*}[ht]
\centering
\includegraphics[width=1.3\columnwidth]{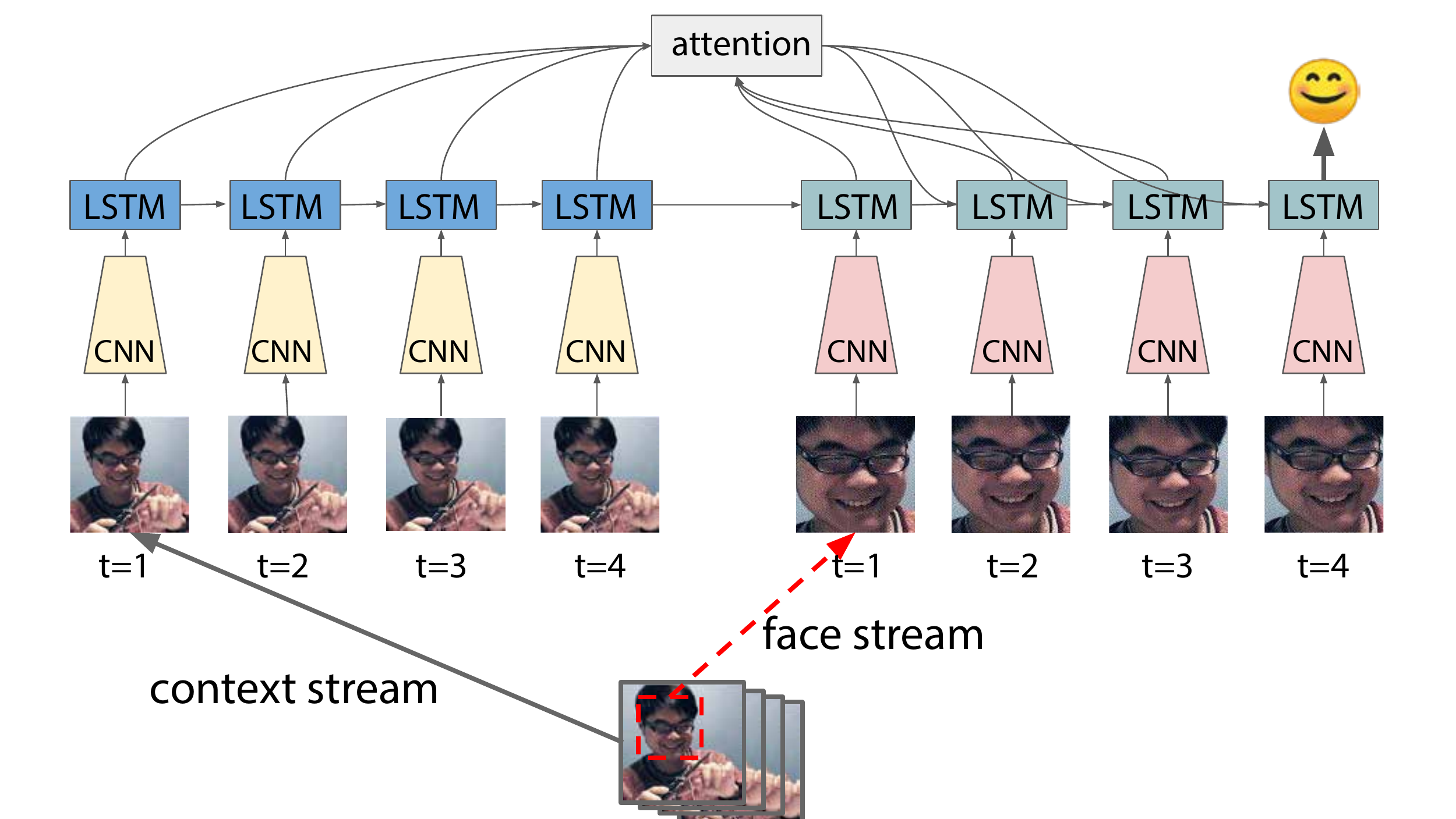}
\caption{Context-Aware Cascade Attention-based RNN (CACA-RNN). A video clip is preprocessed into a face stream (sequence of cropped faces) and a context stream (sequence of original frames). Then, the extracted features of the two sequences are fed into two LSTMs: the context RNN learns contextual temporal information and the face RNN learns facial information. The two LSTMs are in a cascaded architecture with attention mechanism. The context RNN stores context clue in LSTM cells and initiates the first state of face RNN at the first time step. The face RNN then learns from face features and processed context information from attention mechanism. Then the output of face RNN of the last frame of clip is considered to be the prediction of the clip.
}
\label{fig-caca-RNN}
\end{figure*}

\subsection{Attention Mechanism} \label{atten}

Attention mechanism has been proposed to improve the performance of English-to-French machine translation in \cite{Bahdanau2015_nmt} as a sequence-to-sequence model. The mechanism makes decoder not only depend the context vector from encoder but allow the decoder model to learn relevant parts over source sequence to predict target.

The soft dot global attention in \cite{Luong2015_attention} is adapted here. An alignment vector $\va_i = (a_{i, 1}, \ldots, a_{i,T})$ is a sequence over input time, where $a_{i, t}$ is derived by encoder's hidden states $\vh_{e, t}$ at input time $t$ and current decoder's hidden state $\vh_{d, i}$ at output time $i$, which is
\begin{align}
a_{i,t} = \frac{\exp(\score(\vh_{d,i}, \vh_{e,t}))}{\sum_{t=1}^{T}{\exp(\score(\vh_{d,i}, \vh_{e, t}))}}. \label{eq_align}
\end{align}
The score function implemented here is dot operation from $\vh_{d,i}$ and $\vh_{e, t}$, where $\score(\vh_{d, i}, \vh_{e, t}) = \vh_{d,i}^T\vh_{e, t}$ and noted that the hidden size of encoder and decoder should be equal. Then the context vector $\vc_i$ at output time $i$ is derived as weighted sum of all source $\vh_e$ and $a_{i,t}$, which is computed as
\begin{align}
\vc_i = \sum_{t=1}^{T_x}{a_{i,t} \cdot \vh_{e, t}}. \label{eq_c}
\end{align}

To summarize (\ref{eq_align}) and (\ref{eq_c}), the context vector with attention mechanism is a function of hidden states of both encoder and decoder. Based on both decoder and encoder's hidden states, the decoder can pay attention to relevant input sequence of encoder and generate aligned output sequence.

\subsection{Context-Aware Cascade Attention-based RNN} \label{CACA-model}
Context-Aware Cascade Attention-based RNN (CACA-RNN) is proposed to leverage both face and context features to perform emotion recognition. The CACA-RNN consists of two LSTMs in a hierarchical cascade architecture with attention mechanism (Fig. \ref{fig-caca-RNN}).


Unlike encoder-decoder framework, which the decoder is a generative model, there are two encoder-like RNNs in CACA-RNN, ``context RNN'' and ``face RNN'', which they read face feature and context feature respectively. The two RNNs are cascaded with attention mechanism in between. The attention mechanism of the CACA-RNN can locate relevant context information from the context RNN when processing the face sequence in the face RNN. After the context RNN reads whole context features, the face RNN encodes face features and its LSTM context vector is derived from attention operation of face and context RNN's hidden states. Then the face RNN outputs the final prediction class after forwarding whole sequence. 

The context RNN reads the context feature sequence from time step $1$ to $T$ as $\vX_{\text{context}} = (\vx_{\text{context}_1}, \ldots, \vx_{\text{context}_T})$ into a LSTM context vector $\vc_{\text{context}_T}$, where its hidden states at time $t$ being denoted as $\vh_{\text{context}_t} = f(\vx_{\text{context}_t}, \vh_{\text{context}_{t-1}})$. Similarily, the face RNN reads face feature sequences $\vX_{\text{face}} = (\vx_{\text{face}_1}, \ldots, \vx_{\text{face}_T})$ and its conditional probability is modeled as
\begin{align}
p(\vy_i |& \vx_{\text{face}_1}, \ldots, \vx_{\text{face}_i}, \vc_{\text{face}_i}) \nonumber\\
&= h(\vx_{\text{face}_i}, \vc_{\text{face}_i}, \vh_{\text{face}_i}),
\end{align}
where $\vh_{\text{face}_i}$ is the hidden state of the face RNN, denoted as $\vh_{\text{face}_i} = f(\vh_{\text{face}_{i-1}}, \vx_{\text{face}_i}, \vc_{\text{face}_i})$. And the LSTM context vector $\vc_{\text{face}_i}$ is derived from Eq.~\eqref{eq_align} and Eq.~\eqref{eq_c}, which is

\begin{align}
\vc_{\text{face}_i} = \sum_{t=1}^{T}{\frac{\exp(\score(\vh_{\text{face}_i}, \vh_{\text{context}_t}))}{\sum_{t=1}^{T}{\exp(\score(\vh_{\text{face}_i}, \vh_{\text{context}_t}))}} \cdot \vh_{\text{context}_t}}.
\end{align}

\section{Comparison architectures} \label{compar_arch}

\begin{figure}
  \centering
  \begin{subfigure}[b]{0.40\columnwidth}
    \includegraphics[width=\textwidth]{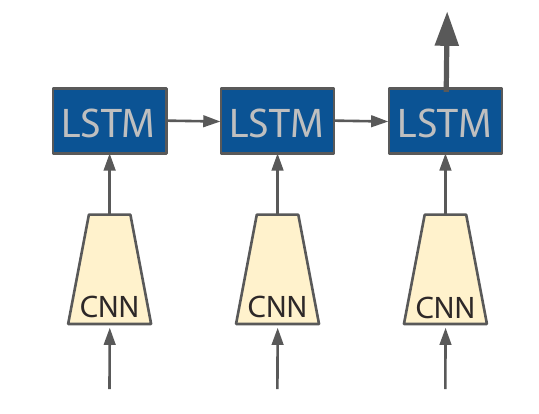}
    \caption{Face-RNN and Context-RNN}
    \label{fig-Face-RNN}
  \end{subfigure}
  \begin{subfigure}[b]{0.48\columnwidth}
    \includegraphics[width=\textwidth]{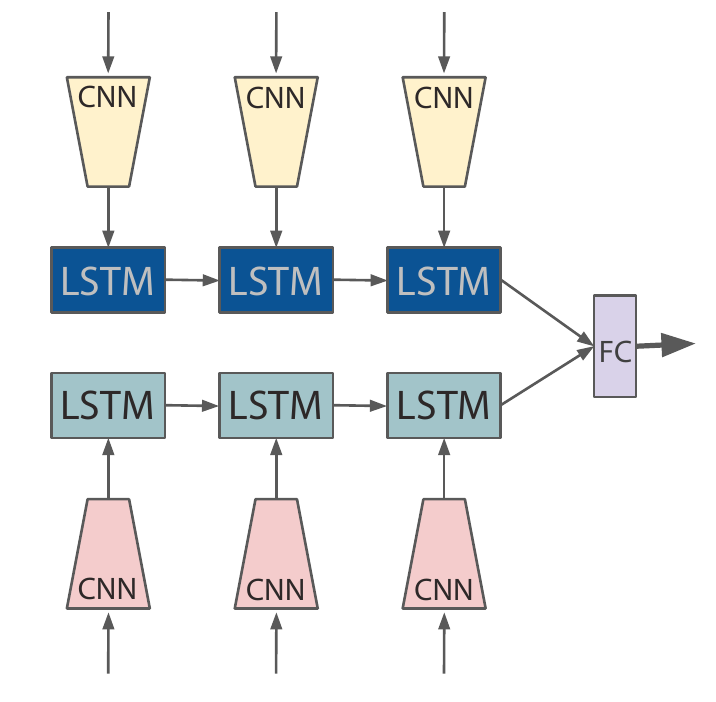}
    \caption{Parallel-RNN}
    \label{fig-parallel-RNN}
  \end{subfigure}
  \begin{subfigure}[b]{0.43\columnwidth}
      \includegraphics[width=\textwidth]{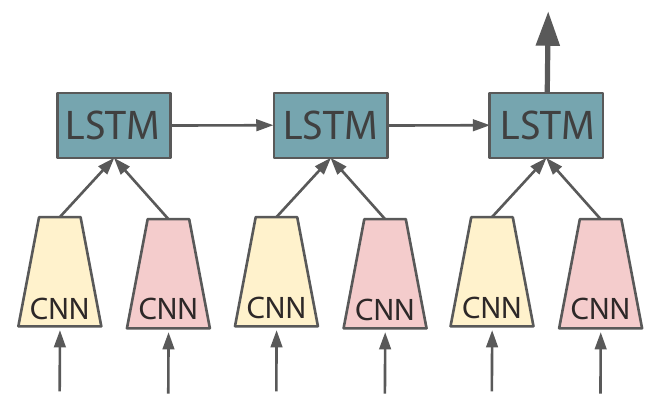}
      \caption{Concatenated-RNN}
      \label{fig-concat-RNN}
  \end{subfigure}
  \begin{subfigure}[b]{0.48\columnwidth}
    \includegraphics[width=\textwidth]{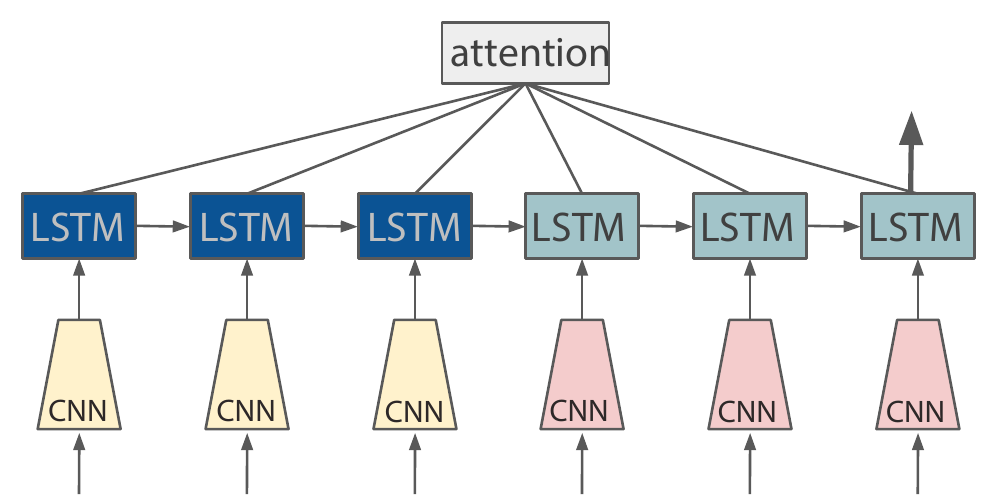}
    \caption{CACA-RNN}
    \label{fig-cascade-RNN}
  \end{subfigure}

\caption{Comparison of RNN-based emotion recognition architectures using face and/or context streams as input. Both streams are processed by a CNN feature extractor into corresponding feature stream as LSTM's input. (a) A single LSTM performs emotion prediction from a feature stream. (b) Two LSTMs reading two streams individually. Hidden states of the two LSTMs are fused together by fully-connected network for final prediction. (c) Two feature streams are concatenated together as input of single LSTM. (d) A cascaded RNN architecture of two LSTMs processing two feature streams: the left LSTM reads one feature stream and store a context vector in its hidden states. And the right LSTM reads the other feature stream with its hidden states is initiated by the left LSTM's context vector. The right LSTM's hidden states are generated from its input and attention from the left LSTM.}
\label{fig-compar}
\end{figure}

To investigate performance of RNN-based models fusing face and context features on emotion recognition, various architectures of combining both features or using only one of the features are introduced as the following:

\textbf{Context-RNN} performs emotion classification using video frames as input, depicted in Fig. \ref{fig-Face-RNN}. The model is similar to a model in \cite{Donahue2014_LTRN} for action recognition, where a LSTM processes sequential features from CNN and learns to predict human action. The Context-RNN reads context features which are extracted from a CNN feature extractor, and last prediction is taken as output. 

\textbf{Face-RNN}, on the contrary to the Context-RNN, is only using face feature sequence to predict emotion. The architecture is depicted in Fig. \ref{fig-Face-RNN}.

\textbf{Parallel-RNN}, illustrated in Fig. \ref{fig-parallel-RNN}, consists of two RNNs processing face and context features individually. Then the hidden states of the two RNNs are fused for final prediction at latest time step.

\textbf{Concatenated-RNN} contains a RNN taking concatenated face and context features as input. Then its output at the last time step is taken to be the prediction of the video clip. (Fig. \ref{fig-concat-RNN})

\textbf{CACA-RNN A}, Context-Aware Cascade Attention-based RNN, consists of two RNN in a cascade architecture with attention mechanism (Fig. \ref{fig-caca-RNN}). The left RNN reads context stream into a context vector. And the right RNN takes attention-processed  context vector and face stream as input, then predicts output at the latest time. 

\textbf{CACA-RNN B} has the same structure as CACA-RNN A (Fig. \ref{fig-caca-RNN}) except for its left RNN reads face stream and the right RNN reads context stream.

\section{Experiments and Evaluation}

\subsection{Datasets}
Chinese Natural Audio-Visual Emotion Database (CHEAVD) 2.0 is used by Multimodal Emotion Recognition Challenge (MEC) 2017 challenge. CHEAVD 2.0 includes 4917 training clips, 707 validation clips and 1406 testing clips from Chinese movies and TV programs. Each clip is labeled with one emotion category according to both video and audio content in 8 classes: ``happy'', ``sad'', ``angry'', ``surprise'', ``disgust'', ``worried'', ``anxious'', and ``neutral''. 

To enlarge the training set, the submitted model was trained with additional private dataset. The dataset includes 4562 video clips (``happy'': 409, ``sad'': 951, ``angry'': 212, ``surprise'': 357, ``disgust'': 270, ``worried'': 58, ``anxious'': 88 , and ``neutral'': 359) and training data clips are trimmed from the videos.

\subsection{Feature Extraction}

For each video clip, two streams are generated in preprocessing stage: ``face stream'' and ``context stream''. Face stream contains sequence of each detected faces from each frame, cropped and scaled to $128\times 128$. With each detected face, corresponding context stream contains the original frame, center-cropped and scaled to $224\times 224$. 

Then, two pre-trained CNNs are used to extract features from the face and context streams as the input of CACA-RNN's face RNN and context RNN. The face feature extractor is a classifier trained with a private dataset for image emotion classification without last layer. And the context feature extractor is a pre-trained model from Squeezenet \cite{Iandola2016_sqz} with ImageNet classification \cite{Russakovsky2015_imagnet}, which the output of the last convolution layer is used as context feature.

\subsection{Training Details}

Videos are downsampled to $5$ fps in both training and inference phases. In training phase, each video is sampled from a random initial frame for data augmentation. In evaluation phase, the initial frame is fixed to be the first frame of video clip. 

For the submitted result, Adam \cite{Kingma2014_adam} was used with learning rate $10^{-4}$ with mini-batch size $32$. The weights of feature extractor CNNs are fixed while training. There is a pooling layer after the context CNN to down sample feature size to size $25088$ as the input of context RNN. Also, a fully connected layer is added after the face CNN to encode face features with vector of size $128$. Additionally, the face RNN is a two-layer LSTM with $128$ hidden states. The context RNN is also a LSTM with $128$ hidden states and attention mechanism is on face RNN.

\begin{table}[htbp]
\begin{center}
\begin{tabular}{c|c|c|c}
\hline
Model & \#params & mAP (\%) & ACC (\%)\\
\hline\hline
Face-RNN & 1.58M& 38.87 & 52.72 \\
\hline
Context-RNN & 1.19M& 28.74 & 37.68 \\
\hline
Parallel-RNN& 0.86M& 40.44 & 54.30\\
\hline
Concatenated-RNN& 1.38M& 39.78 & 54.15\\
\hline
CACA-RNN A& 0.79M& \textbf{40.73} & 53.01\\
\hline
CACA-RNN B& 0.79M& 39.91 & \textbf{54.44}\\
\hline
\end{tabular}
\caption{Experimental results on MEC2017 validation set.}
\label{tab_compar}
\end{center}
\end{table}

\subsection{Quantitative results}

To explore performance of the architectures described in Section. \ref{compar_arch}, the validation results are summarized in Table. \ref{tab_compar}, with \textit{only} CHEAVD 2.0 was used for training and validation in the experiments.

To make comparison fair, the models are adjusted such that number of parameters are close. For each experiment, training was repeated five times with different random seeds and median performance metrics are selected among them. The feature sizes of face feature and context feature are encoded to $128$ by fully-connected layers.
Face-RNN, Context-RNN and Concatenated-RNN has a two-layer LSTM with $256$ hidden nodes. In Parallel-RNN, there are two two-layer LSTMs of $128$ hidden nodes. And the first LSTM of CACA-RNN also has two layers with $128$ hidden states, the second RNN is a LSTM with $128$ hidden nodes.

Notably, the models fusing both context and face clues outperform the models using only one of the features. It shows that the context information is helpful but not enough to perform emotion recognition without cropped face.

For models fusing face and context features, the results show that Parallel-RNN outperforms Concatenated-RNN. Learning from multiple different feature sequences can be more effective for networks with multiple branches than a single branch model.
The performance of CACA-RNN A and CACA-RNN B is higher than the others in evaluation in mAP (mean average precision) and accuracy. It shows that cascade attention-based RNN can improve the performance of handling two kinds of features.

\section{Results on MEC 2017}

\begin{figure}[h]
  \centering
  \begin{subfigure}{0.9\columnwidth}
    \includegraphics[width=\textwidth]{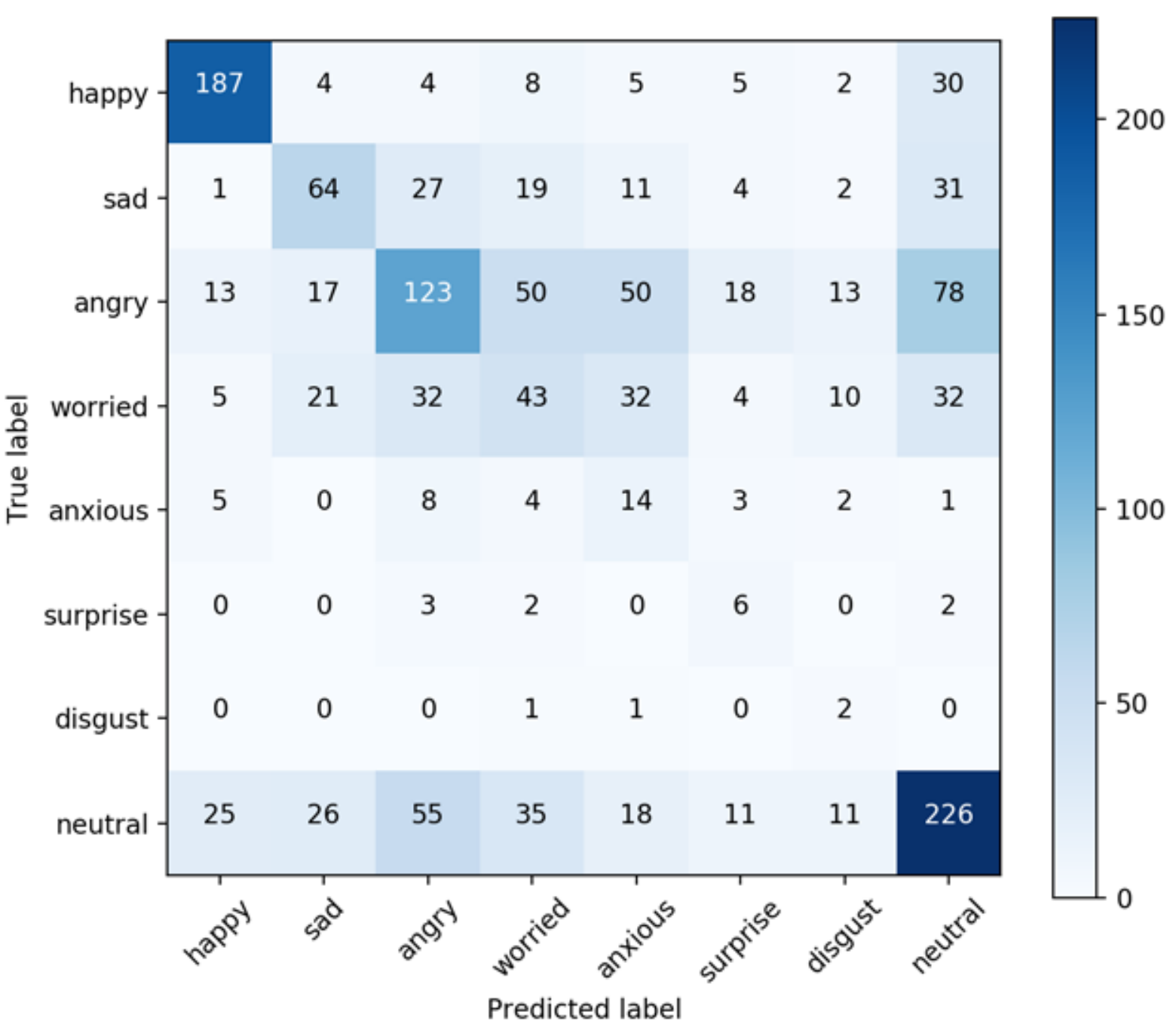}
    \caption{CACA-RNN}
    \label{fig-cm-CACA}
  \end{subfigure}\\
  \begin{subfigure}{0.9\columnwidth}
    \includegraphics[width=\textwidth]{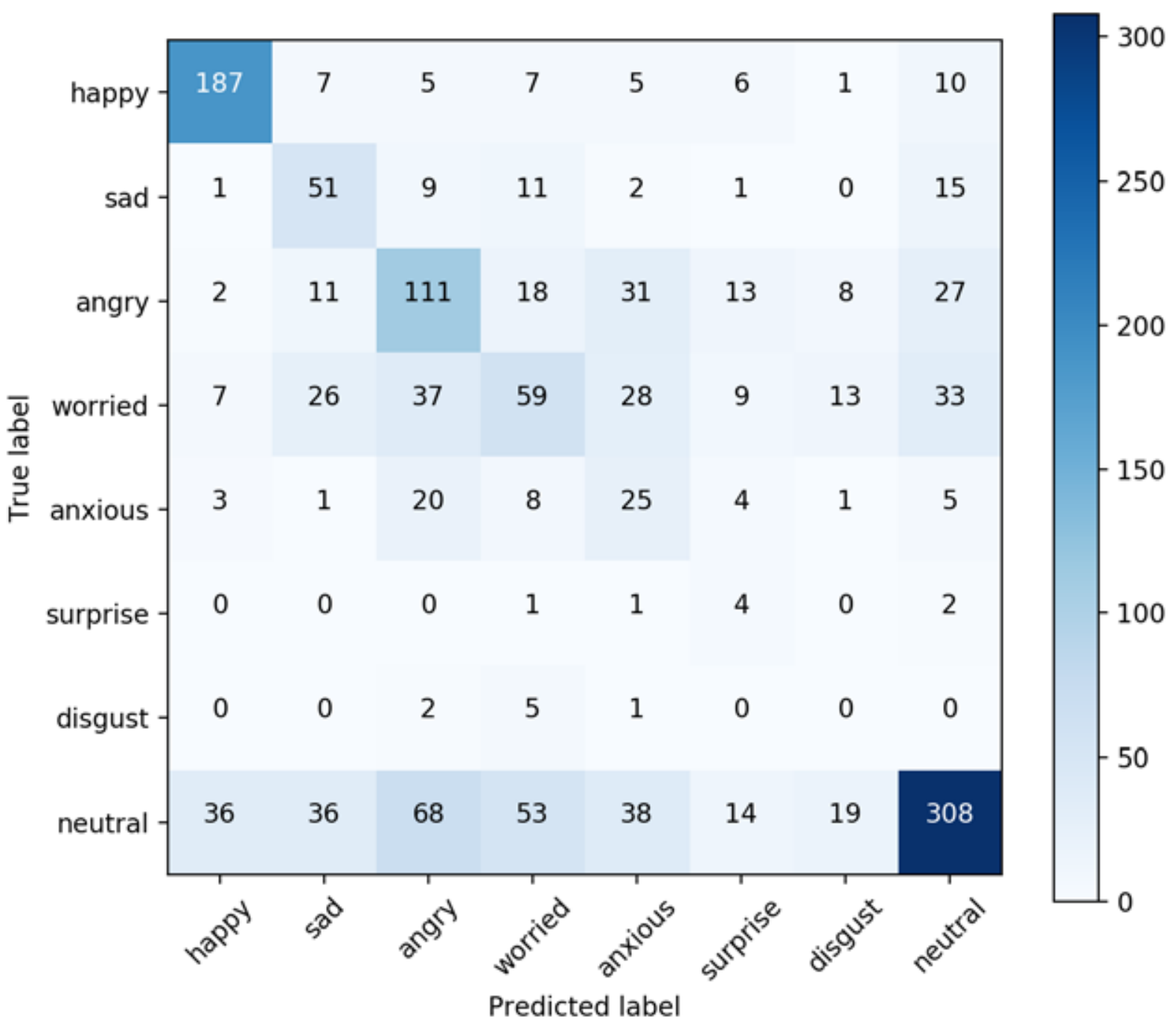}
    \caption{CACA-RNN+2D-3D-CNN}
    \label{fig-cm-ensemble}
  \end{subfigure}
\caption{Confusion matrices of MEC2017 testing set. (a) The model has the highest mAP among five submissions in MEC2017.  (b) The model has the highest accuracy among five submissions in MEC2017.}
\label{cm}
\end{figure}

Five submissions are allowed to submit to video-based emotion recognition sub-challenge in MEC2017. Evaluation on mean average precision (mAP) and accuracy are both considered in MEC, two (of five) evaluation results of submissions are shown in Table. \ref{tab_mec}: (1) ``CACA-RNN'', described in previous sections, the one with highest mAP, (2) ``CACA-RNN+2D-3D-CNN'', the one with highest best accuracy. Their confusion matrices are shown in Fig. \ref{cm}.

\begin{figure}[htb]
\centering
\includegraphics[width=0.8\columnwidth]{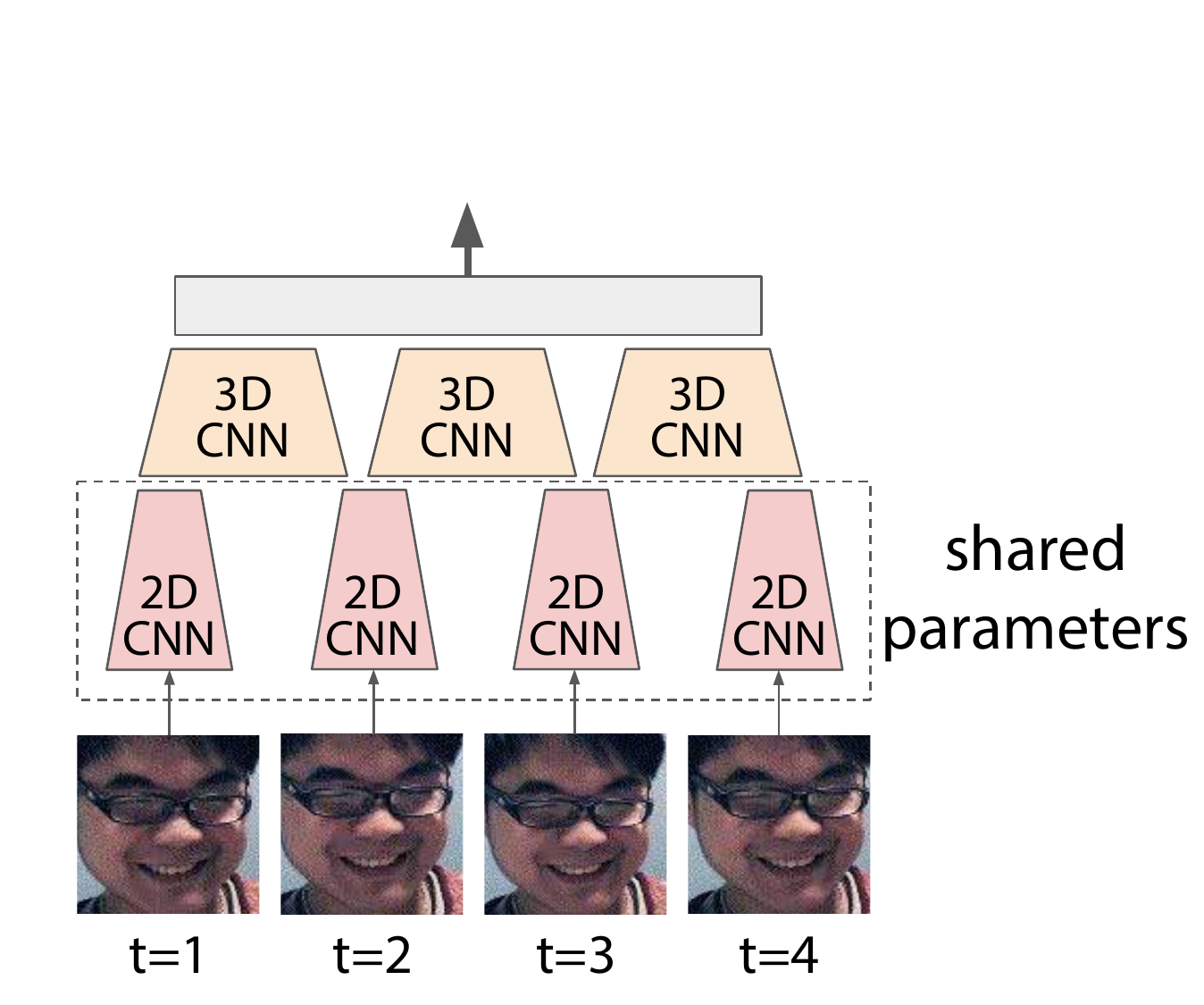}
\caption{A 2D-3D-CNN model for emotion recognition. The video stream is preprocessed into face stream of cropped faces. Each face of face stream is first encoded by a pre-trained CNN face feature extractor to high level feature maps. Then a 3D convolution neural network learns spatiaotemporal kernel from the feature map stream. There is an adaptive pooling before classifier to handle various length of input.}
\label{fig-3dcnn}
\end{figure}

CACA-RNN+2D-3D-CNN is an ensemble network of CACA-RNN and a 2D-3D-CNN network using linear weighted output from each model as the output prediction. The 2D-3D-CNN network is a spatiotemporal convolutional network, depicted in Fig. \ref{fig-3dcnn}, learning temporal information from high level face feature maps from the same face feature extractor CNN in CACA-RNN.

\begin{table}[htbp]
\begin{center}
\begin{tabular}{c|c c|c c}
\hline
\multirow{2}{*}{Method} & \multicolumn{2}{c|}{Validation Set} & \multicolumn{2}{c}{Testing Set} \\
 & mAP (\%) & ACC (\%) & mAP (\%) & ACC (\%)\\
\hline\hline
CACA-RNN & 41.53 & 51.34 & \textbf{45.51} &  47.30 \\
\hline
\pbox{5cm}{CACA-RNN+\\2D-3D-CNN} & 52.16 & 56.58 & 44.74 &  \textbf{52.99} \\
\hline
baseline \cite{Li2017_mec} & 34.1 & 36.5 & 21.7 & 35.3 \\
\hline
\end{tabular}
\end{center}
\caption{Submitted Results to MEC2017.}
\label{tab_mec}
\end{table}

\newpage

\section{Conclusion} 
In this work, context information is first shown to be helpful in emotion perception. Then, different architectures using facial and/or context information for video emotion recognition was implemented and evaluation results were compared. It shows that the models using both information can achieve the highest accuracy.

Among them, a novel architecture, CACA-RNN, was proposed with a cascaded LSTM attention-based architecture to leverage both face and context information from video. CACA-RNN has the best performance in the compared models. CACA-RNN consists of two LSTMs, context RNN and face RNN, processing context and face features respectively, and attention mechanism in the face RNN enables it to learn relationship to the context RNN and fuse information from two sequences. Compared to a multi-branch RNN (Parallel-RNN) and a single RNN handling concatenated features (Concatenated-RNN), CACA-RNN outperforms on evaluation MEC2017 validation dataset. The experiments also shows that context information improves for video emotion recognition. The models with additional context features perform better than the models using only face features.

CACA-RNN may be further extended to fusing more kinds of inputs for multi-modal emotion recognition or be extended to other video-based tasks such as action recognition.


\end{document}